\documentclass{article}

\usepackage[nonatbib,final]{neurips_2020}

\usepackage[utf8]{inputenc} 
\usepackage[T1]{fontenc}    
\usepackage{hyperref}       
\usepackage{url}            
\usepackage{booktabs}       
\usepackage{amsfonts}       
\usepackage{nicefrac}       
\usepackage{microtype}      
\usepackage[pdftex]{graphicx}
\usepackage{amsmath}
\usepackage{chngcntr}

\title{Generating Synthetic Multispectral Satellite Imagery from Sentinel-2}

\author{
  Tharun Mohandoss \\
  Radiant Earth Foundation\\
  San Francisco, CA 94105 \\
  \texttt{tharun96@utexas.edu} \\
     \And
   Aditya Kulkarni \\
   Radiant Earth Foundation\\
  San Francisco, CA 94105 \\
   \texttt{adkulkar@eng.ucsd.edu} \\
   \AND
   Daniel Northrup \\
   Benson Hill \\
  Saint Louis, MI \\
   \texttt{dan@northrup.ag} \\
   \And
   Ernest Mwebaze \\
   Sunbird AI \\
  Kampala, Uganda \\
   \texttt{emwebaze@gmail.com} 
   \And
  Hamed Alemohammad \\
   Radiant Earth Foundation\\
  San Francisco, CA 94105 \\
  \texttt{hamed@radiant.earth} \\
}

\begin{document}
\maketitle
\begin{abstract}
Multi-spectral satellite imagery provides valuable data at global scale for many environmental and socio-economic applications. Building supervised machine learning models based on these imagery, however, may require ground reference labels which are not available at global scale. Here, we propose a generative model to produce multi-resolution multi-spectral imagery based on Sentinel-2 data. The resulting synthetic images are indistinguishable from real ones by humans. This technique paves the road for future work to generate labeled synthetic imagery that can be used for data augmentation in data scarce regions and applications. 

\end{abstract}

\section{Introduction}

In many supervised learning problems with satellite imagery, such as agricultural monitoring, training data are generated from data collected on the ground (a.k.a. ground reference). Ground reference data collection is an extensive effort, and extremely scarce in remote and dangerous areas that would most benefit from remote sensing applications. As a result, there are limited number of training datasets available for these problems which narrows application of machine learning (ML) based techniques with satellite imagery to specific parts of the world while these images are available at global scale. 

Techniques such as transfer learning \cite{10.1145/3209811.3212707} and data augmentation 
\cite{doi:10.1080/15481603.2017.1323377} have been used to mitigate this issue. Transfer learning has shown some promising results for crop type classification in regions with homogeneous agricultural practices such as the US~\cite{Wang2019}. However, this is not necessarily scalable to other parts of the world, such as regions that are dominated by smallholder farming. Similar study for a road detection problem shows that a model trained in Las Vegas, US is not easily transferable to Khartoum, Sudan~\cite{Nachmany_2019}.

Several studies have used GANs with satellite imagery for applications such as super-resolution and data augmentation~\cite{8059820, DBLP:journals/corr/abs-1809-04985, Jiang2019}. These show the success of applying GAN to high-resolution satellite imagery with 3-4 multispectral bands.

GANs are generally used to reproduce an RGB image distribution but remote sensing data augmentation applications would require producing images with more than 3 bands along with their corresponding labels. Generalizing from RGB imagery to labelled multi-band imagery can be achieved in two steps by first producing more than 3 bands and then incorporating label generation.  Here, we tackle the first step and present a new framework for generating synthetic satellite imagery. We introduce a new GAN architecture that incorporates ten bands with varying resolutions from Sentinel-2 satellite imagery, and generates realistic multispectral imagery for data augmentation. Multispectral bands are essential for many land surface monitoring applications, therefore this model can be applied to augment data across multiple applications.  

\section{Dataset}
We use multispectral imagery from Sentinel-2 satellites (details of the individual bands are listed in the Table~\ref{s2bands}). Our study region is western Kenya, where we have ground reference data for the next phase of this project. Therefore, for the experiments of this study we selected 105k Sentinel-2 images from this region at $256\times256$ pixels to be used as training images. The selected images consist of several types of land cover including barren land, urban areas, forests, croplands, and water. We filter out samples that contain cloud and samples where water bodies cover a significant portion of the image. Figure~\ref{fig:main_experiments} shows sample real RGB images used in this study, and Figure~\ref{fig:true_10_band_image} shows all ten bands from sample images.

\section{Methodology}
GANs suffer from a variety of issues such as training instability and mode collapse that require careful setting of hyperparameters for successful training. ~\cite{DBLP:journals/corr/GulrajaniAADC17} suggests WGAN-GP training loss which can mitigate the issues caused by insufficient overlap between generated and target distributions leading to an increased stability of training. Karnewar et al.~\cite{DBLP:journals/corr/abs-1903-06048} builds on the previous work ~\cite{DBLP:journals/corr/abs-1710-10196} and proposes the Multi-Scale Gradients (MSG) method where images are produced in exponentially growing resolutions starting from the smallest size of $4 \times 4$ till the final output size allowing the discriminator to provide feedback in all resolutions instead of just one which further mitigates these issues.

\subsection{Model}
We adopt the MSG-GAN architecture and WGAN-GP loss. Specifically, we use the MSG-ProGAN architecture used in~\cite{DBLP:journals/corr/abs-1903-06048} (and variants of it) and modify it to produce multispectral satellite images of size $256 \times 256$ instead of RGB images (which differs in the number of output channels). The exact architectures are shown in Tables \ref{table_gen} and \ref{table_disc}. 

\begin{figure}
    \centering
    \includegraphics[width=0.8\linewidth]{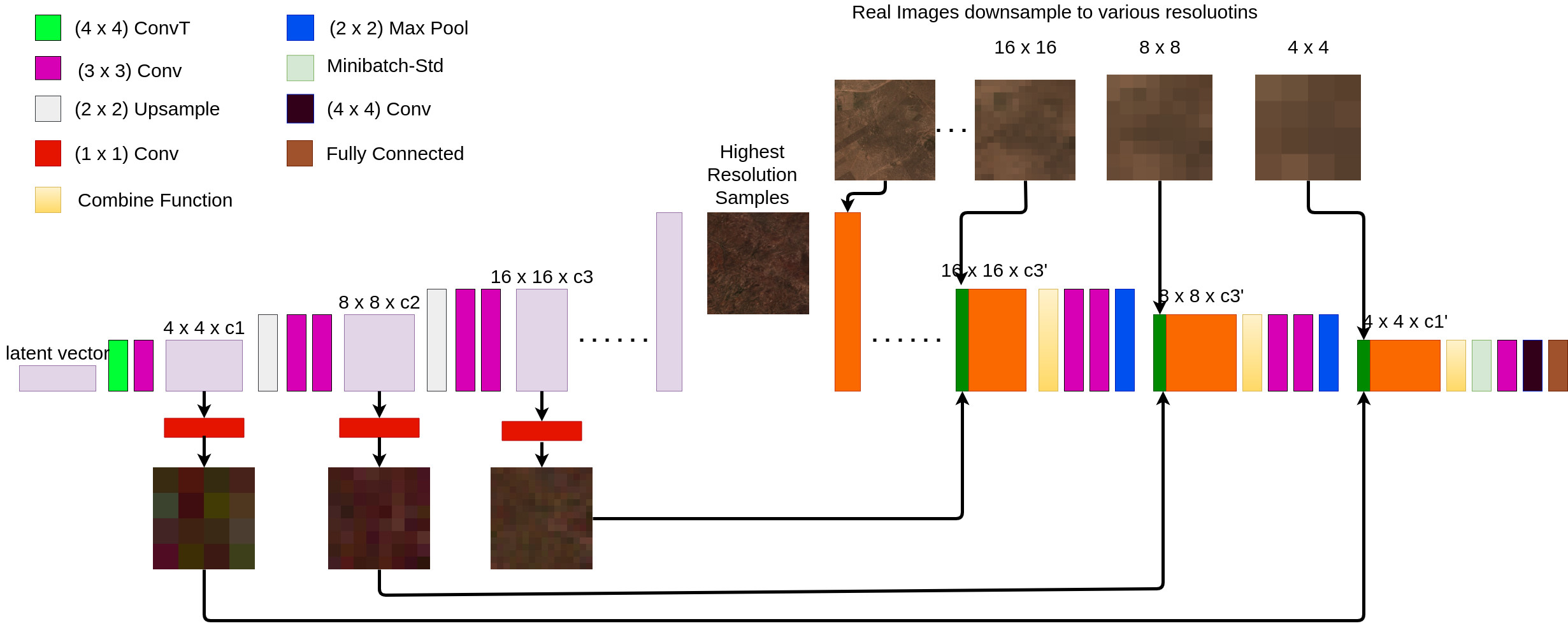}
    \caption{Block Diagram of the MSG-ProGAN model}
    \label{fig:MSG-ProGAN}
\end{figure}

\section{Experiments}
Satellite imagery has coarser pixel size (a.k.a. resolution) compared to common imagery in computer vision, and they usually have more than 3 bands, possibly with varying resolutions. Due to these differences, a straightforward port of GAN models from existing literature might not achieve the desired results. Therefore, we broke down our approach to smaller experiments and incrementally increased complexity of the model towards the final goal of producing multispectral satellite images. 

We start with experiment 1 where we generate $4\times4$ resized versions of the RGB bands of the target images using the smaller parts of the Generator an Discriminator in the original MSG-ProGAN network.  Results are satisfactory (not shown) and the distribution of the $4\times4$ images produced by the generator is visually indistinguishable from the $4\times4$ versions of the original images. Using this as a toy experiment, we chose RMSProp as the optimizer, $10^{-3}$ as the learning rate and a weight factor ($\lambda$) of 10 for the gradient penalty for the rest of experiments. More details on the training is provided in section\ref{training_details}.

Table~\ref{experimentstable} shows the list of all experiments conducted. We start with $256\times256$ RGB imagery and then generate $256\times256$ ten band imagery. Finally, we experiment with smaller versions of the MSG-ProGAN model to optimize the model resource footprint (E.g. GPU memory requirement).  

\begin{table}
    \caption{List of experiments}
    \label{experimentstable}
    \centering
    \begin{tabular}{cp{5.5cm}p{6.6cm}} 
        \toprule
        No & Experiment & Result \\
        \midrule
        1 & $4\times4$ RGB & Synthetic and real images are indistinguishable.\\
        2 & $256\times256$ RGB & Synthetic images can fool humans.\\
        3A & $256\times256$ ten bands 1 & Less than satisfactory results.\\
        3B & $256\times256$ ten bands 2 & Less than satisfactory results.\\
        3C & $256\times256$ ten bands 3 & Best results for the ten bands case.\\
        4A & $256\times256$ RGB fewer convolutions & Satisfactory Results.\\
        4B & $256\times256$ RGB fewer filters & Image quality almost as good as Experiment 2.\\
        5 & $256\times256$ ten bands fewer filters & Image quality almost as good as Experiment 3C.\\
        \bottomrule
    \end{tabular}
\end{table}

\subsection{Experiment 2: Generating 256$\times$256 RGB images}
Here, we use the full MSG-ProGAN architecture for $256\times256$ images described in Tables~\ref{table_gen} and~\ref{table_disc} by setting m = 3 to produce a 3 channel RGB output. We train the model for 12 epochs on the RGB channels with a batch size of 4. The resulting synthetic images are extremely convincing. In an online experiment with 106 unique individuals attempting to distinguish real images from synthetic ones, a majority of the them (>70) scored an average in the range of 50\%-70\%. Despite the GAN having learned well, there are some classes that are missing in the outputs such as urban areas and regular croplands. This could be due to the class imbalance in our dataset (regular croplands and urban areas are underrepresented). Figure~\ref{fig:main_experiments} shows example synthetic images from this experiment.

\subsection{Experiment 3: Generating 256$\times$256 ten band images}
Here, we expand the model to ten bands consisting of four bands of 10m resolution and six bands of 20m resolution. The 10m bands form a $256\times256$ image and the 20m bands form a $128\times128$ image due to the difference in resolutions. In experiment 3A, we simply modify the model from experiment 2 and use nearest neighbor to interpolate the $128\times128$ images into $256\times256$ images and treat all the ten bands the same. Next, we set $m=10$ in the architecture of MSG-ProGAN to produce ten bands per image instead of three. The results are less than satisfactory. Some examples shown in Figure \ref{fig:Experiment_appendix}.

\begin{figure}
    \centering
    \includegraphics[width=.9\linewidth]{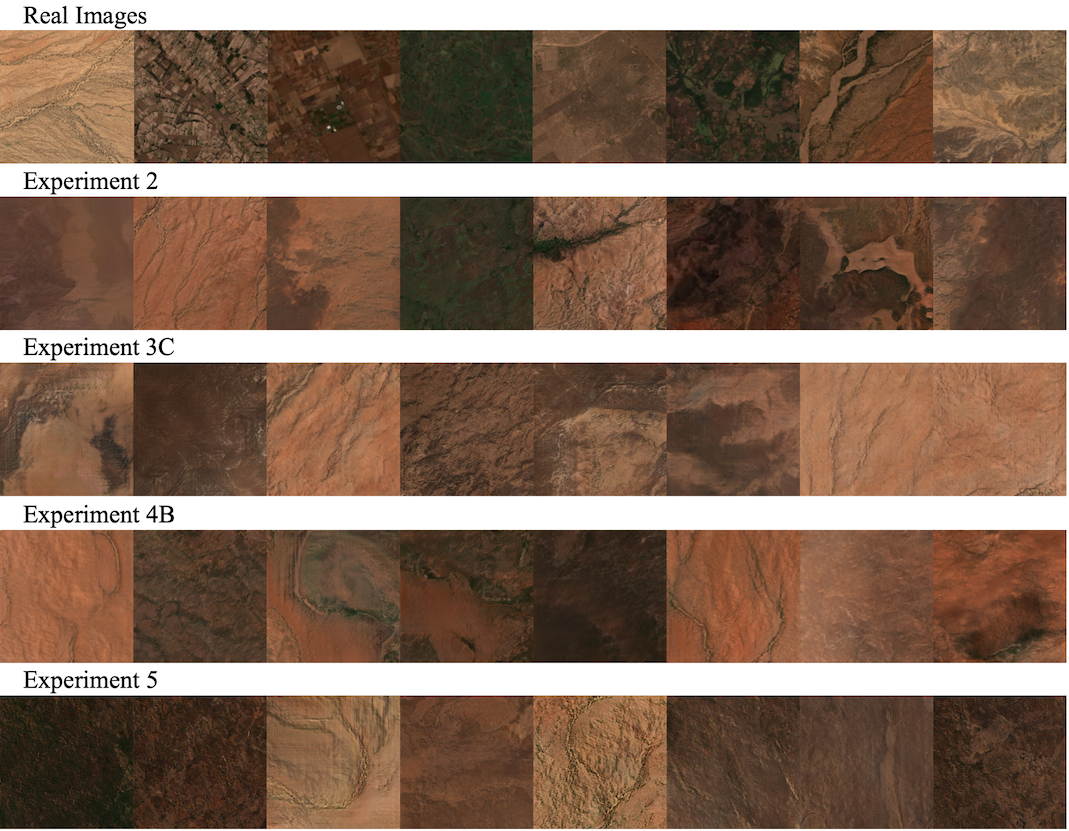}
    \caption{Sample real and synthetic images from experiments 2, 3C, 4B, and 5 (only RGB bands shown)}
    \label{fig:main_experiments}
\end{figure}

For experiment 3B, we grouped together bands based on their resolutions (forming 2 groups, one for 10m bands, one for 20m bands). In the generator, we produce $256\times256$ output for the 10m bands as usual and $128\times128$ output for the 20m bands by using an extra pooling layer after the convolution layer that produces the image for this group. In the discriminator, we process these groups in parallel branches and merge the output features together forming a neural network described in Figure~\ref{fig:3B_model}. The images produced by this architecture are of poor quality and suffer from various issues such as presence of artefacts and loss of correlation between the bands of different resolutions in the generated images. Some results (only the RGB bands) are shown in Figure~\ref{fig:Experiment_appendix}.

To improve the results and ensure bands with varying resolution are correlated, we revised the model from 3A to generate only the 10m bands in the highest resolution ($256\times256$). The proposed discriminator model for this is presented in Figure~\ref{fig:3C_model}. This architecture produces the best quality images out of the three experiments (3A, 3B, and 3C) designed to generate ten band images. The images from different bands are well synced and the images are almost realistic although there is some observable loss of variety in the images. Figure~\ref{fig:main_experiments} shows samples generated from this model. 
    
\subsection{Experiments 4A, 4B, and 5: Resource efficient models}
To reduce the resource footprint of our model, we designed two experiments for RGB images. First, we investigate the effect of reducing the number of convolutions in the model by removing every other convolution and also reducing the number of images produced by the generator and passed to the discriminator (only $4\times4$, $16\times16$, $64\times64$, $256\times256$ ones as shown in Figure~\ref{fig:MSG-ProGAN_fewer_conv}). Next, we investigate the effect of reducing the number of filters in every layer of the model by half.

We find that both of these approaches are viable strategies to reduce the model's resource footprint i.e they both produce satisfactory results but reducing the number of filters is the preferred approach and the performance in this case is very close to the original model. We also observe by modifying the model from Experiment 3C in the same way that this result generalizes to the ten bands case.

\section{Discussion}
Satellite imagery have unique properties that distinguishes them from common images used in the computer vision field. As a result, existing GAN architectures cannot be applied to these imagery off the shelf. With the goal of generating synthetic satellite imagery for data augmentation, we designed multiple GAN architectures of varying complexity to generate synthetic multispectral multi-resolution satellite imagery. Results show:
\vspace{-0.15cm}
\begin{enumerate}
    \item A straightforward port of models from existing GAN literature is sufficient to generate RGB bands of satellite images but these models have high resource requirements.
    \vspace{-0.15cm}
    \item Our modified GAN architecture can generate realistic images with ten bands at varying resolutions.
    \vspace{-0.15cm}
    \item Significant GAN model size reduction can be achieved by using fewer number of filters with minimal reduction in quality.
\end{enumerate}
\vspace{-0.15cm}
This study demonstrates a state-of-the-art GAN architecture that can be used to generate diverse multispectral satellite imagery with varying band resolutions. Such a model is an essential step to the next phase of our experiment which is generating these images with their corresponding labels.

\section*{Acknowledgements}
This research is funded by a grant awarded to Radiant Earth Foundation through the 2019 Grand Challenges Annual Meeting Call-to-Action from the Bill \& Melinda Gates Foundation. The findings and conclusions contained within are those of the authors and do not necessarily reflect positions or policies of the Bill \& Melinda Gates Foundation.

\bibliographystyle{unsrt}
\bibliography{S2_Synthetic_TD}

\pagebreak

\appendix
\counterwithin{figure}{section}
\counterwithin{table}{section}

\section{Supplementary Figures and Tables}

\begin{table}[h]
    \caption{Description of the bands in Sentinel-2 satellite images. 10m and 20m bands are used in this study.}
    \label{s2bands}
    \centering
    \begin{tabular}{cc} 
        \toprule
        Band & Spatial Resolution \\
        \cmidrule(r){1-2}
        Band 1 – Coastal aerosol     	&	60\\
        Band 2 – Blue	                &	10\\
        Band 3 – Green	                &	10\\
        Band 4 – Red	                &	10\\
        Band 5 – Vegetation red edge 1 	&   20\\
        Band 6 – Vegetation red edge 2  &	20\\
        Band 7 – Vegetation red edge 3  &   20\\
        Band 8 – NIR                    &	10\\
        Band 8A – Narrow NIR            &	20\\
        Band 9 – Water vapour           &	60\\
        Band 10 – SWIR - Cirrus	        &	60\\
        Band 11 – SWIR1	       	        &   20\\
        Band 12 – SWIR2                 &   20\\
     \bottomrule
    \end{tabular}
\end{table}

\begin{table}[h]
   \caption{Generator Architecture. After every block in the generator, a $1\times1$ Conv layer is used to convert the output activation volume into an image which is passed onto the discriminator. }
    \label{table_gen}
    \centering
    \begin{tabular}{cccc} 
     \toprule
     Block & Operation & Activation & Output Shape \\ 
     \midrule
        & Latent Vector & Norm  & 512 x 1 x 1 \\
     1. & Conv 4x4      & LReLU & 512 x 4 x 4  \\ 
        & Conv 3x3      & LReLU & 512 x 4 x 4  \\
        & Upsample      & -     & 512 x 8 x 8  \\
     2. & Conv 3x3      & LReLU & 512 x 8 x 8  \\ 
        & Conv 3x3      & LReLU & 512 x 8 x 8  \\ 
        & Upsample      & -     & 512 x 16 x 16 \\
     3. & Conv 3x3      & LReLU & 512 x 16 x 16  \\ 
        & Conv 3x3      & LReLU & 512 x 16 x 16  \\ 
        & Upsample      & -     & 512 x 32 x 32 \\
     4. & Conv 3x3      & LReLU & 512 x 32 x 32  \\ 
        & Conv 3x3      & LReLU & 512 x 32 x 32  \\ 
        & Upsample      & -     & 512 x 64 x 64 \\
     5. & Conv 3x3      & LReLU & 256 x 64 x 64  \\ 
        & Conv 3x3      & LReLU & 256 x 64 x 64  \\ 
        & Upsample      & -     & 256 x 128 x 128 \\
     6. & Conv 3x3      & LReLU & 128 x 128 x 128  \\ 
        & Conv 3x3      & LReLU & 128 x 128 x 128  \\ 
        & Upsample      & -     & 128 x 256 x 256 \\
     7. & Conv 3x3      & LReLU & 64 x 256 x 256  \\ 
        & Conv 3x3      & LReLU & 64 x 256 x 256  \\ 
    \bottomrule
    \end{tabular}
\end{table}

\begin{table}
    \caption{Discriminator Architecture. $m$ is the number of output channels which varies across different experiments.}
    \label{table_disc}
    \centering
    \begin{tabular}{cccc} 
     \toprule
     Block & Operation & Activation & Output Shape \\ 
     \midrule
        & Image 1       & -     & m x 256 x 256 \\
        & From RGB      & -     & 64 x 256 x 256  \\ 
     1. & MiniBatchStd  & -     & 65 x 256 x 256  \\
        & Conv 3x3      & LReLU & 64 x 256 x 256  \\
        & Conv 3x3      & LReLU & 128 x 256 x 256  \\
        & Max Pool      & -     & 128 x 128 x 128  \\
        & Image 2       & -     & m x 128 x 128 \\
        & Concat        & -     & (128+m) x 128 x 128  \\ 
     2. & MiniBatchStd  & -     & (128+m+1) x 128 x 128  \\
        & Conv 3x3      & LReLU & 128 x 128 x 128  \\
        & Conv 3x3      & LReLU & 256 x 128 x 128  \\
        & Max Pool      & -     & 256 x 64 x 64  \\
        & Image 3       & -     & m x 64 x 64 \\
        & Concat        & -     & (256+m) x 64 x 64  \\ 
     3. & MiniBatchStd  & -     & (256+m+1) x 64 x 64  \\
        & Conv 3x3      & LReLU & 256 x 64 x 64  \\
        & Conv 3x3      & LReLU & 512 x 64 x 64  \\
        & Max Pool      & -     & 512 x 32 x 32  \\
        & Image 4       & -     & m x 32 x 32 \\
        & Concat        & -     & (512+m) x 32 x 32  \\ 
     4. & MiniBatchStd  & -     & (512+m+1) x 32 x 32  \\
        & Conv 3x3      & LReLU & 512 x 32 x 32  \\
        & Conv 3x3      & LReLU & 512 x 32 x 32  \\
        & Max Pool      & -     & 512 x 16 x 16  \\
        & Image 4       & -     & m x 16 x 16 \\
        & Concat        & -     & (512+m) x 16 x 16  \\ 
     5. & MiniBatchStd  & -     & (512+m+1) x 16 x 16  \\
        & Conv 3x3      & LReLU & 512 x 16 x 16  \\
        & Conv 3x3      & LReLU & 512 x 16 x 16  \\
        & Max Pool      & -     & 512 x 8 x 8  \\
        & Image 5       & -     & m x 8 x 8 \\
        & Concat        & -     & (512+m) x 8 x 8  \\ 
     6. & MiniBatchStd  & -     & (512+m+1) x 8 x 8  \\
        & Conv 3x3      & LReLU & 512 x 8 x 8  \\
        & Conv 3x3      & LReLU & 512 x 8 x 8  \\
        & Max Pool      & -     & 512 x 4 x 4  \\
        & Image 6         & -     & m x 4 x 4 \\
        & Concat          & -     & (512+m) x 4 x 4  \\ 
     7. & MiniBatchStd    & -     & (512+m+1) x 4 x 4  \\
        & Conv 3x3        & LReLU & 512 x 4 x 4  \\
        & Conv 3x3        & LReLU & 512 x 4 x 4  \\
        & Fully Connected & Linear     & 1 x 1 x 1  \\
     \bottomrule
    \end{tabular}

\end{table}

\begin{figure}
    \centering
    \includegraphics[width=\linewidth]{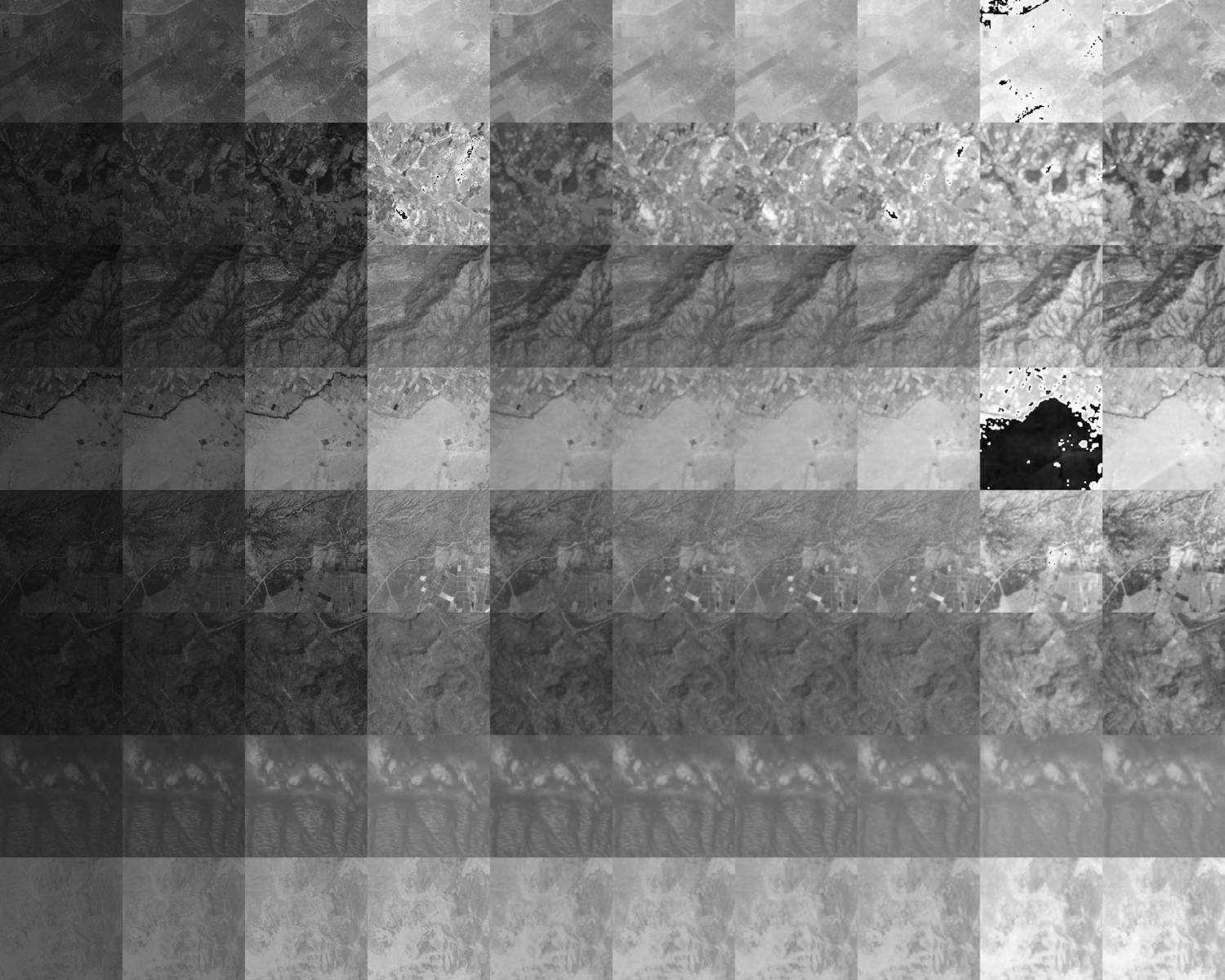}
    \caption{Samples of real ten band images in our dataset: Each row shows one sample and each column is a band from left to right: blue, green, red, NIR , red-edge, red-edge-2, 'red-edge-3', 'red-edge-4', 'SWIR1', 'SWIR2'. The first 4 bands are 10m bands ($256\times256$) and the remaining six are 20m bands ($128\times128$). The 20m bands have been resized to $256\times256$ for ease of visualization.}
    \label{fig:true_10_band_image}
\end{figure}

\begin{figure}
    \centering
    \includegraphics[width=1\linewidth]{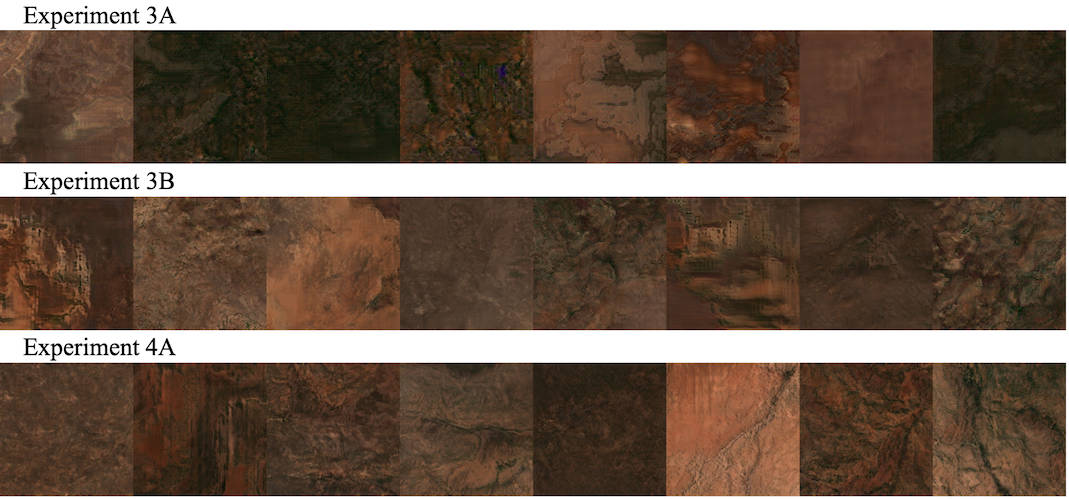}
    \caption{Sample synthetic images from experiments 3A, 3B and 4A (only RGB bands shown)}
    \label{fig:Experiment_appendix}
\end{figure}

\begin{figure}
    \centering
    \includegraphics[width=\linewidth]{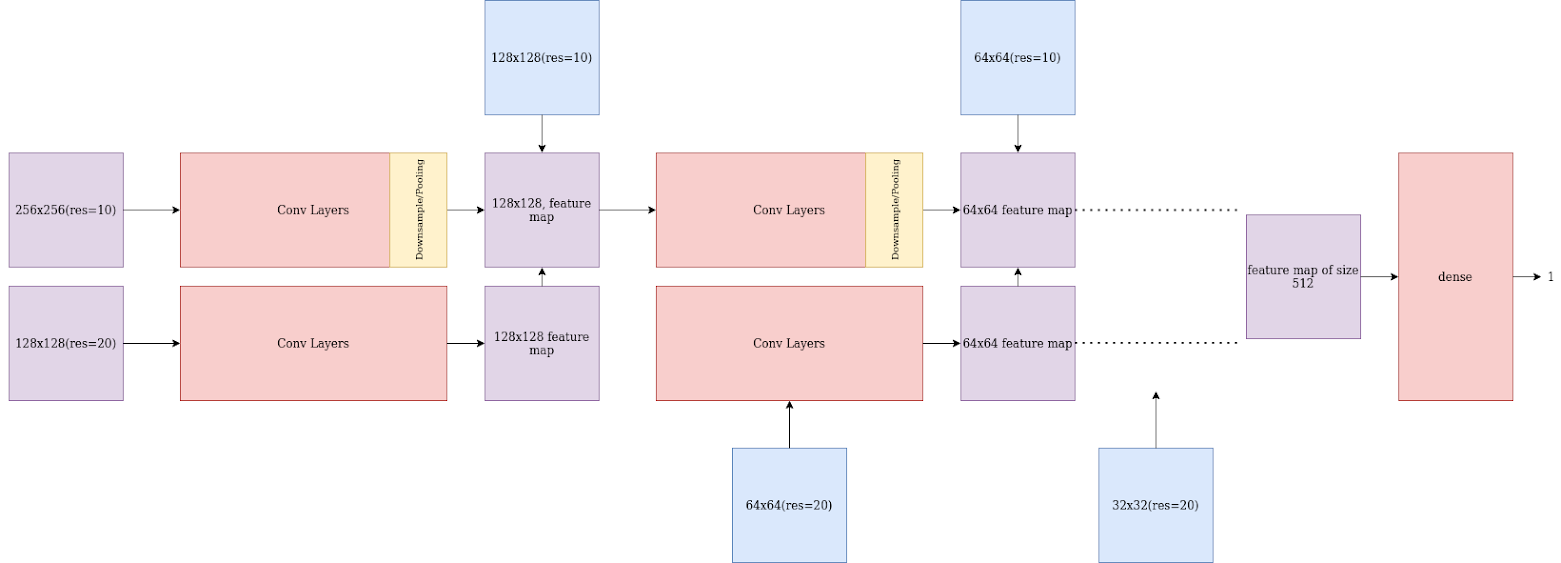}
    \caption{Discriminator model architecture for experiment 3B}
    \label{fig:3B_model}
\end{figure}

\begin{figure}
    \centering
    \includegraphics[width=\linewidth]{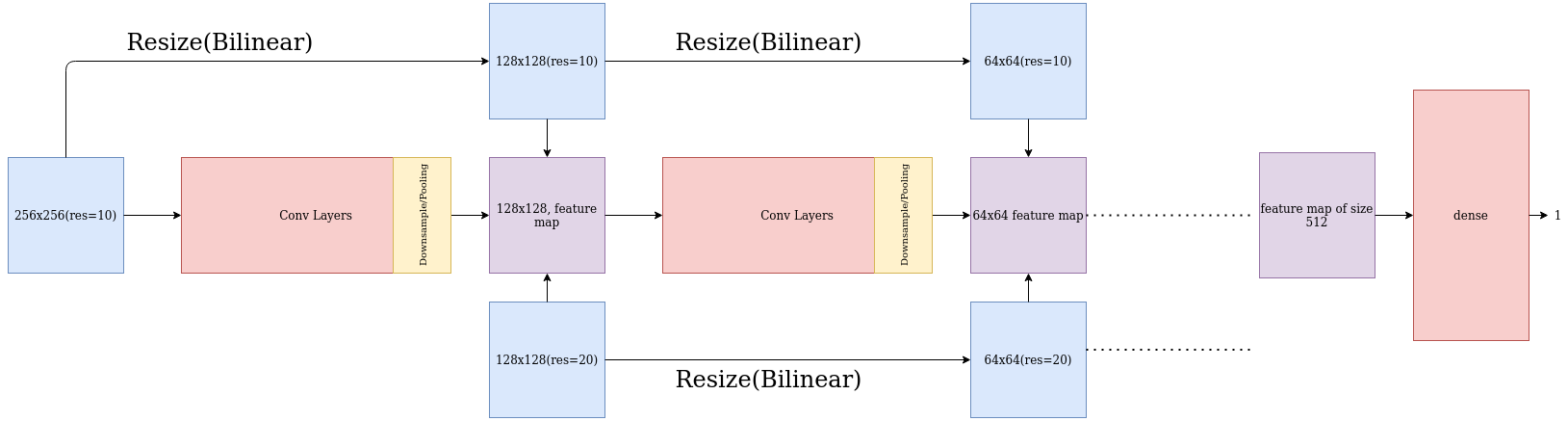}
    \caption{Discriminator model architecture for experiment 3C}
    \label{fig:3C_model}
\end{figure}

\begin{figure}
    \centering
    \includegraphics[width=0.8\linewidth]{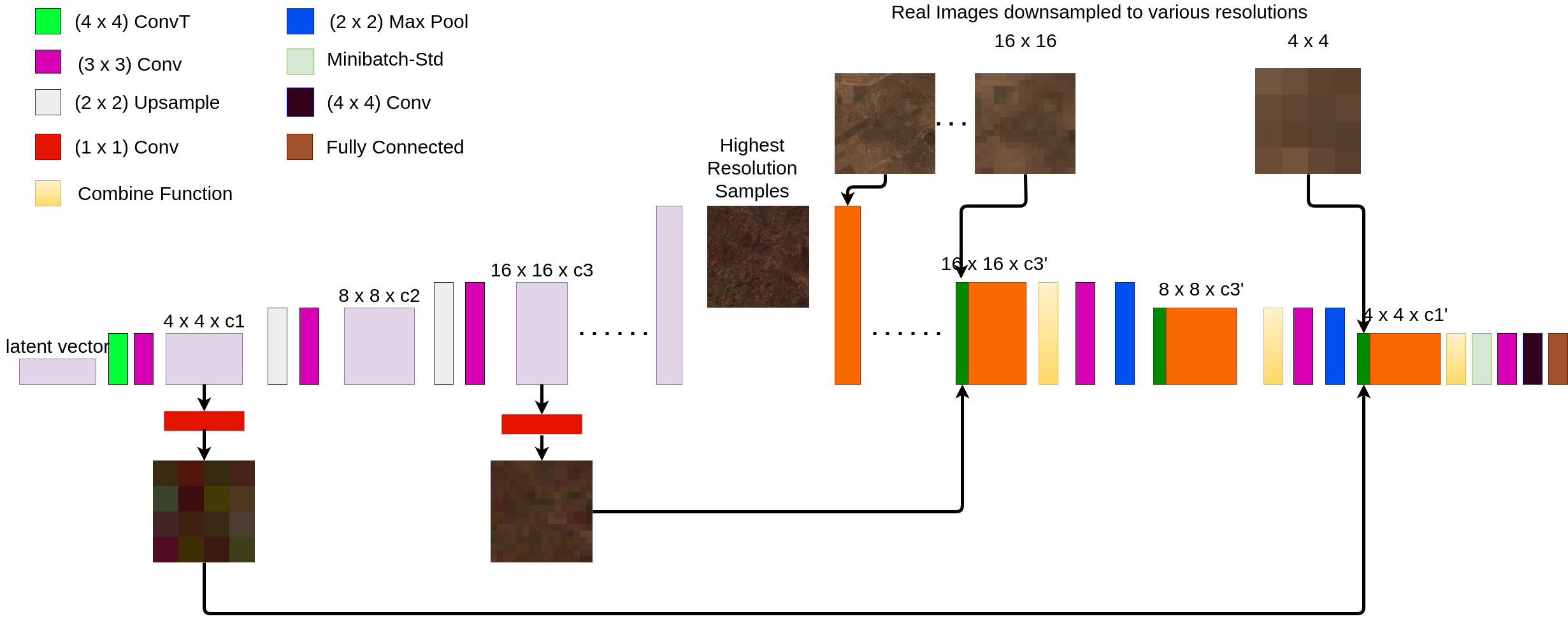}
    \caption{Block Diagram of the modified MSG-ProGAN model with fewer convolutions.}
    \label{fig:MSG-ProGAN_fewer_conv}
\end{figure}

\begin{figure}
    \centering
    \includegraphics[width=\linewidth]{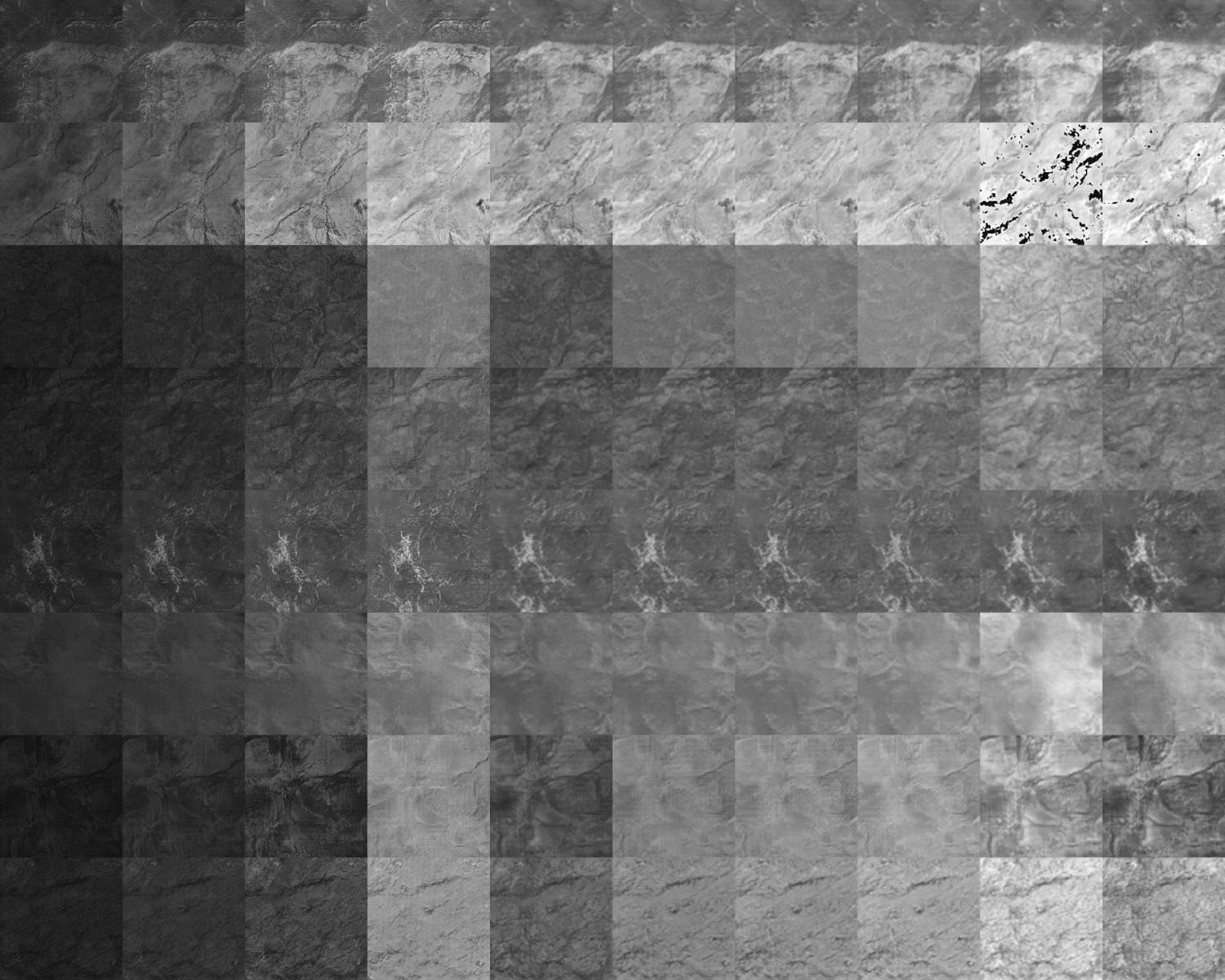}
    \caption{Results of experiment 3C: Each row shows one sample and each column is a band from left to right: blue, green, red, NIR , red-edge, red-edge-2, 'red-edge-3', 'red-edge-4', 'SWIR1', 'SWIR2'. The first 4 bands are 10m bands ($256\times256$) and the remaining six are 20m bands ($128\times128$). The 20m bands have been resized to $256\times256$ for ease of visualization.}
    \label{fig:3C_10_band_image}
\end{figure}

\section{Model and Training Details}
\label{training_details}
\subsection{Weight initialization: Equalized Learning Rate}
We follow the weight initialization technique mentioned in~\cite{DBLP:journals/corr/abs-1710-10196} to use a trivial N (0, 1)
initialization and then explicitly scale the weights at run-time~\cite{DBLP:journals/corr/HeZR015}. We set $w'_i$ = $w_i/c$,
where $w_i$ are the weights and c is the per-layer normalization constant from He's initializer. This ensures that the dynamic range, and thus the learning speed, is the same for all weights.

\subsection{Pixel-wise Feature Vector Normalization in Generator}
We also apply the layer-wise vector normalization in~\cite{DBLP:journals/corr/abs-1710-10196} after every 3x3 convolution in the generator to disallow the scenario where the magnitudes in the generator and discriminator spiral out of control as a result of unhealthy competition between the discriminator and generator.

\subsection{Minibatch standard deviation}
Introduced first in the~\cite{NIPS2016_6125} minibatch discrimination can help GANs prevent mode collapse. We follow~\cite{DBLP:journals/corr/abs-1903-06048} in their use of minibatch standard deviation as a simpler variant of the above technique to achieve the same goal. We first compute the standard deviation for each feature in each spatial location over the minibatch. We then average these estimates over all features and spatial locations to arrive at a single value. We replicate the value and concatenate it to all spatial locations and over the minibatch, yielding one additional (constant) feature map. This additional layer is added after every concatenation step. 

\subsection{Training Details}
We use Pytorch's RMSProp optimizer with a learning rate of $10^{-3}$ for both the generator and the discriminator. The total loss for the generator and discriminator are as follows where D(x) represents the discriminator output for input image x :
\begin{equation}
\begin{split}
    G_{loss} &= - D(generated_{image}) \\
    D_{loss} &= D(generated_{image}) - D(real_{images}) + \lambda \times GP
\end{split}
\end{equation}

where GP denotes the Gradient Penalty corresponding to the WGAN-GP loss. We set the relative importance parameter $\lambda$ to 10.

\end{document}